\documentclass{article}
\usepackage{spconf,amsmath,graphicx}
\usepackage{times}
\usepackage{latexsym}
\usepackage{algorithm}
\usepackage{algorithmic}
\usepackage{footnote}
\usepackage{url}
\usepackage{multirow}
\usepackage[switch]{lineno}
\usepackage{booktabs}
\usepackage{makecell}
\usepackage{amsfonts}
\usepackage{amsmath}
\usepackage{caption}
\usepackage{subcaption}
\usepackage{CJKutf8}
\usepackage{makecell}

\usepackage[utf8]{inputenc}
\usepackage{enumitem}
\usepackage{microtype}
\usepackage{graphicx}
\usepackage{amsmath}
\usepackage{bbm}

\newcommand{\fref}[1]{Fig.~\ref{#1}}
\newcommand{\tref}[1]{Table~\ref{#1}}
\newcommand{\sref}[1]{\S\ref{#1}}
\newtheorem{definition}{Definition}

\title{Rethinking Targeted Adversarial Attacks for Neural Machine Translation}
\name{Junjie Wu$^{\star }$ \qquad Lemao Liu$^{\dagger}$ \qquad Wei Bi$^{\dagger}$ \qquad Dit-Yan Yeung$^{\star }$}
  
  \address{$^{\star}$ \textit{HKUST}, Hong Kong \qquad
      $^{\dagger}$ \textit{Tencent AI Lab}}

\begin{document}
\begin{CJK*}{UTF8}{gbsn}
\ninept
\maketitle
\begin{abstract}
Targeted adversarial attacks are widely used to evaluate the robustness of neural machine translation systems. Unfortunately, this paper first identifies a critical issue in the existing settings of NMT targeted adversarial attacks, where %our empirical quantification indicates that 
their attacking results are largely overestimated. To this end, this paper presents a new setting for NMT targeted adversarial attacks that could lead to reliable attacking results. Under the new setting, it then proposes a \textbf{T}argeted \textbf{W}ord \textbf{G}radient adversarial \textbf{A}ttack (\textbf{TWGA}) method to 
craft adversarial examples.
%for evaluation. 
%Extensive results show that two prevailing NMT systems are vulnerable to %these targeted adversarial examples, while TWGA can effectively generate these adversarial examples.
%targeted adversarial examples crafted under our proposed setting, while TWGA can effectively generate such adversarial examples.
Experimental results demonstrate %on two prevailing NMT systems show 
that our proposed setting could provide faithful attacking results for %could reliably evaluate the performances of 
targeted adversarial attacks on NMT systems, and the proposed TWGA method can effectively attack such victim NMT systems. In-depth analyses on a large-scale dataset further illustrate some valuable findings.~\footnote{This research has been made possible by funding support from the Research Grants Council of Hong Kong under the General Research Fund project 16204720 and the Research Impact Fund project R6003-21.} Our code and data are available at \url{https://github.com/wujunjie1998/TWGA}.
\end{abstract}
\section{Introduction}
\label{introduction}
Neural machine translation~(NMT)~\cite{vaswani2017attention} has been the prevailing framework for machine translation in recent years. However, current NMT systems are still fragile
to inputs with imperceptible perturbations, also known as adversarial examples,
%in that even imperceptible noises in the source side 
in that their translation qualities could be dramatically degraded~\cite{belinkov2018synthetic,michel2019evaluation,zou2019reinforced,lai2022generating}. 
%Therefore, %studies on robustness of NMT is a hot research topic recently and 
%various methods have been proposed to evaluate the adversarial robustness, i.e., the robustness against adversarial examples, of NMT systems,
%~\cite{michel2019evaluation,zou2019reinforced,cheng2020seq2sick,zhang2021crafting,lai2022generating}.
%In existing works, adversarial robustness are measured by
%Various adversarial attack methods have been proposed to generate adversarial examples for NMT systems, which can be mainly split into two categories. %, i.e., sentence-level attack and word-level targeted attack. 
%In existing works, adversarial robustness are measured by two main categories.
%The first category~\cite{belinkov2018synthetic,zhang2021crafting,wan2022paeg,lai2022generating,sadrizadeh2023transfool} %analyzes the degradation of the translation of an adversarial input sentence 
%crafts adversarial examples to degrade the translation quality of NMT models
%in terms of a sentence-level metric such as BLEU~\cite{papineni2002bleu}. %The adversarial input may slightly differ from the original input at any positions. 
%On the other hand, the second category~-~%which is the focus of this paper, 
Among these attack methods, targeted adversarial attack that enforces targeted translation errors on some targeted words rather than the entire sentence
%incorrect words in the translation of an adversarial input
%\footnote{We name this category NMT targeted adversarial attack in the rest of the paper.} 
has drew many research attention~\cite{ebrahimi2018adversarial, cheng2020seq2sick, wallace2020imitation, sadrizadeh2023targeted} since % . These targeted attacks are 
it can provide more fine-grained attacking results.% regarding the instability of NMT systems.

\begin{figure}
    \centering
    \includegraphics[width=0.44\textwidth]{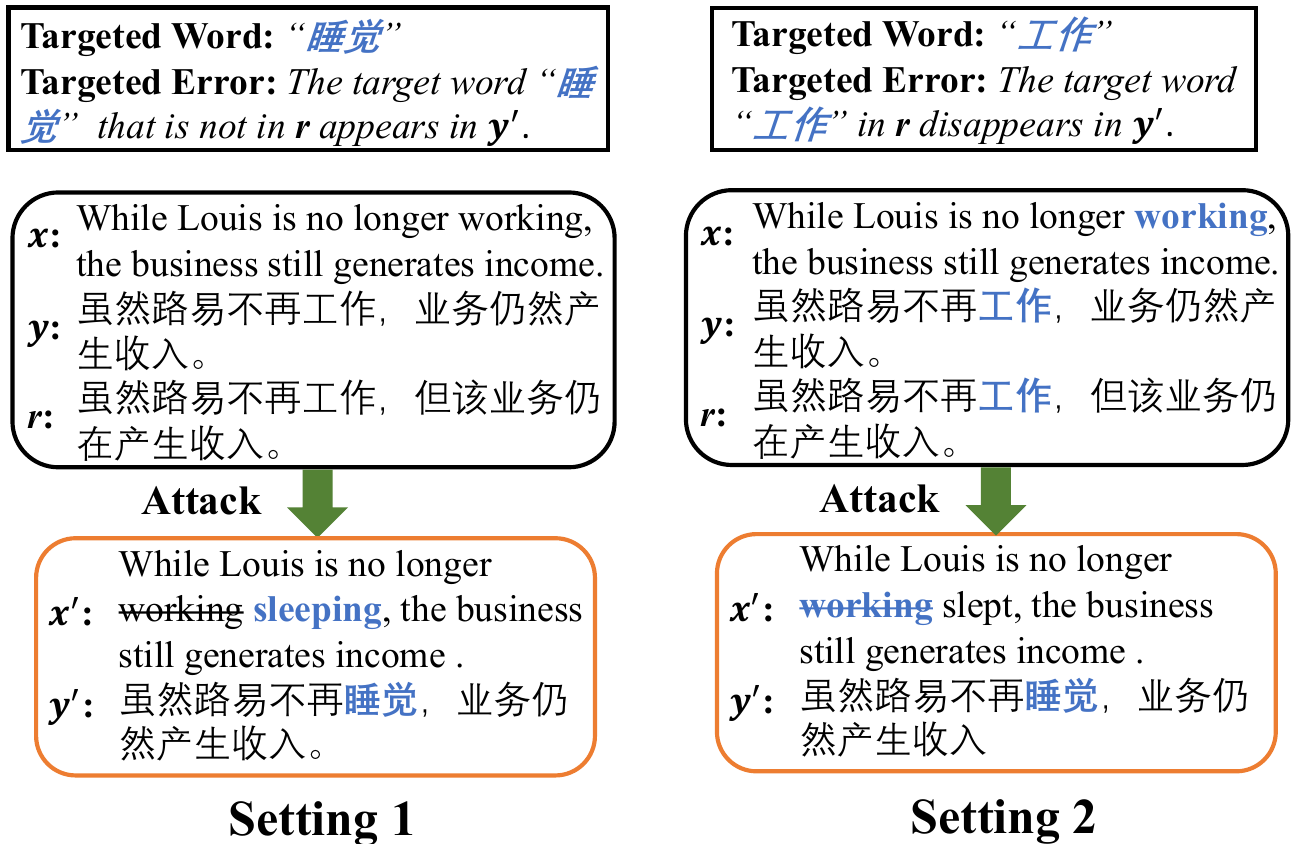}
    \caption{Illustration of invalid adversarial examples in existing settings of targeted adversarial attacks on an NMT system. %$\boldsymbol{x}$ is the original input and 
    %$\boldsymbol{x'}$ is the attacked source,
    %%and $\boldsymbol{y'}$ are the output of $\boldsymbol{x}$ and $\boldsymbol{x'}$ from the NMT system, $\boldsymbol{r}$ is the reference of $\boldsymbol{x}$. %We use the attack proposed by~\cite{cheng2020seq2sick} and~\cite{wallace2020imitation} for Setting 1 and Setting 2, respectively. %Target words and their translations are colored with blue. Modifications are colored with red.
    %Both adversarial examples are actually invalid because \textit{``working"} in the source sentence is crafted into \textit{``sleeping"} (in setting 1) or \textit{``slept"} (in setting 2), whose ground-truth translation is not ``工作'' any more but ``睡觉''. } 
    }
    \label{fig:compare settings}
   \vspace{-1.5em}
\end{figure}

Based on the types of targeted translation errors, we first frame the problem settings of existing NMT targeted adversarial attacks as: make a targeted word that is not in the original reference appear in the translation output~\cite{cheng2020seq2sick, sadrizadeh2023targeted} (\textbf{Setting 1}); %while (setting 2) 
or let a targeted word in the original reference  %in the reference translation 
disappear in the translation output~\cite{ebrahimi2018adversarial, wallace2020imitation} (\textbf{Setting 2}). The evaluation of targeted attacks under both settings %usually assume that a crafted adversarial example keep 
relies on a label-preserving assumption, i.e., an adversarial example keeps the same semantic meaning of the original input sentence and thus it is rational to verify the effectiveness of this adversarial example using the reference of the original sentence, which is further illustrated in \sref{revisiting}. However, we argue that this assumption is not always true since existing settings do not constrain the source tokens that can be perturbed, which makes many crafted adversarial examples invalid as shown in \fref{fig:compare settings}. 
%For example, in~\fref{fig:compare settings}, since the source word ``working" in the source sentence $\boldsymbol{x}$ is perturbed into ``sleeping" or ``slept", its ground-truth translation is not ``工作" any more but ``睡觉",  both adversarial examples are invalid. 
Hence, the attack success rates obtained under both problem settings are largely overestimated (see the results in \tref{tab:empirical quantification}), making the obtained attacking results of those targeted attacks unreliable.

To solve the above issue, in this paper, we first propose a new setting for NMT targeted adversarial attacks such that the obtained attacking results are faithful. Specifically, given a source sentence including a targeted word, %the translation correctness of the target word can be simply measured by whether its translations appear in the translation output.
an attack method is applied to make 
%the goal of an attack is to make 
the correct translation of this targeted word disappear in the output by only modifying non-targeted word tokens in the source sentence, which ensures the validity of the crafted adversarial example. In addition,
since a targeted word might have various meanings under different contexts, we introduce a bilingual dictionary to ensure that all possible reference translations of this targeted word are not shown in the translation output. Moreover, since the semantic meaning of a targeted word could be destroyed in a meaningless source sentence with messy contents, we require that adversarial examples crafted under our setting should be fluent and meaningful.  %, which further enhancing the reliability of our setting. 
Noting that no sentence-level references are needed during the attack process, making our setting easily to be scaled up to yield more unbiased results. 

Next, under this new attack setup, we design a white-box \textbf{T}argeted \textbf{W}ord \textbf{G}radient adversarial \textbf{A}ttack (\textbf{TWGA}) method to generate %bilingual adversarial pairs. 
adversarial examples for NMT systems.
%an evaluated NMT system. The adversarial robustness of this NMT model is finally measured on the crafted adversarial examples.
On two translation datasets, we use several NMT targeted adversarial attacks including TWGA to 
%evaluate the adversarial robustness of 
attack two popular NMT systems under our proposed setup, and the results illustrate that TWGA can effectively craft high-quality adversarial examples. Additionally, unlike existing %targeted attacks
works that mainly implement NMT targeted adversarial attacks on small datasets, we conduct experiments on a large-scale dataset for the first time and make in-depth analyses of the attacking results, leading to some insightful findings.  

\iffalse
Overall, our main contributions are as follows:
%\begin{enumerate}
\begin{enumerate}[wide=0\parindent,noitemsep,topsep=0em]
%\setlength{\itemsep}{0.5pt}
%\setlength{\parsep}{0.5pt}
%\setlength{\parskip}{0.5pt}
    \item We point out a critical limitation in existing settings for the research works on NMT targeted adversarial attacks, i.e., unreliable attacking results.  
    %(\lm{Reviewers may argue that bilingual dictionary is built via maual annotations.} either manually annotated or automatically generated).
    \item We propose a new setting for NMT targeted adversarial attacks that could yield more reliable attacking results %for NMT systems 
    on the translation of any interested targeted words. 
    \item Under the new setting, we design a novel NMT targeted adversarial attack. Experimental results on both a common-scale and a large-scale dataset show the effectiveness of our proposed attack and illustrate some insightful findings.
\end{enumerate}
\fi

\section{Revisiting Existing Attack Settings}
\label{revisiting}

%\subsection{Existing Settings: Review}
%\label{previous settings}

%\textbf{Existing Settings.} \quad
%\lemao{Please mention the referene translation $\mathbf{r}$.}

\vspace{-0.2cm}
\subsection{Existing Settings: Review}
\vspace{-0.2cm}

\label{previous settings}
We start by describing the two settings mentioned in \sref{introduction}. %and how to find an invalid adversarial example under both settings. 
Consider a source sentence $\boldsymbol{x}$ with a reference $\boldsymbol{r}$ %$\boldsymbol{x} = (x_1, x_2, ..., x_m)$ 
and its translation $\boldsymbol{y}$
%$\boldsymbol{y} = (y_1, y_2, ..., y_n)$ 
generated by an NMT system $\mathcal{M}$. We define $\boldsymbol{x'}$ as a slightly perturbed version of $\boldsymbol{x}$ and $\boldsymbol{y'}$ as its corresponding translation output from $\mathcal{M}$, respectively. 
Let $w$ denote the targeted word. Such $\boldsymbol{x'}$ is an adversarial example in two settings as follows:
%For each input $\boldsymbol{x}$, the goals of targeted attacks under both settings can be summarized as crafting a $\boldsymbol{x'}$ to:
\begin{itemize}[wide=0\parindent,noitemsep,topsep=0em]
    \item \textbf{Setting 1:} $w$ is sampled from the target vocabulary and not included in $\boldsymbol{r}$. %to appear in $\boldsymbol{y'}$~\cite{cheng2020seq2sick}. 
    $\boldsymbol{x'}$ is an adversarial example of $\boldsymbol{x}$ if it leads to a targeted translation error: $w$ appears in $\boldsymbol{y'}$ ~\cite{cheng2020seq2sick,sadrizadeh2023targeted}. 
    \item \textbf{Setting 2:} %make a target word 
    $w$ is a word 
    in $\boldsymbol{r}$. $\boldsymbol{x'}$ is an adversarial example of $\boldsymbol{x}$ if it leads to a targeted translation error: $w$ disappears in $\boldsymbol{y'}$~\cite{ebrahimi2018adversarial,wallace2020imitation}.
    %to disappear in $\boldsymbol{y'}$~\cite{ebrahimi2018adversarial,wallace2020imitation}. %$\boldsymbol{x'}$ is an adversarial example of $\boldsymbol{x}$ if %it make 
    %$w$ disappear in $\boldsymbol{y'}$
\end{itemize}

Both settings above actually make a \textbf{label-preserving assumption}: the perturbed $\boldsymbol{x'}$ keeps the same semantic meaning of $\boldsymbol{x}$, and thus it is rational to verify targeted translation errors in $\boldsymbol{y'}$ according to the reference $\boldsymbol{r}$ of $\boldsymbol{x}$ without figuring out the reference of $\boldsymbol{x'}$. As a result, the left attack in \fref{fig:compare settings} is regarded as successful under setting 1 since $\boldsymbol{y'}$ makes a targeted translation error on the targeted word $w=$``睡觉'' with respect to $\boldsymbol{r}$, and the right attack in \fref{fig:compare settings} is regarded as successful under setting 2 because $\boldsymbol{y'}$ makes a targeted translation error on the targeted word $w=$``工作'' with respect to $\boldsymbol{r}$.

\textbf{Issue.} \quad
\label{issue}
However, as pointed out in \sref{introduction}, %existing settings assume that the perturbations on $\boldsymbol{x}$ should be small enough to ensure the similarity between $\boldsymbol{x}$ and its adversarial example $\boldsymbol{x'}$. 
the label-preserving assumption made by both settings 1\&2 is not always true since both settings do not constrain the tokens that can be perturbed. 
%This assumption enables they can examine whether translation errors on target words by comparing $\boldsymbol{y'}$ and $\boldsymbol{x}$'s translation $\boldsymbol{y}$ outputted by $\mathcal{M}$.
Consider the two examples $\boldsymbol{x'}$ in \fref{fig:compare settings} again. Under setting 1, the targeted word $w=$``睡觉'' appears in $\boldsymbol{y'}$ while its translation \textit{``sleeping''} is also added to $\boldsymbol{x'}$. As for setting 2, the targeted word $w=$``工作'' disappears in $\boldsymbol{y'}$ while its source translation \textit{``working''} is also deleted in $\boldsymbol{x'}$. %In other words, 
According to humans, ``睡觉'' in both $\boldsymbol{y'}$ is the ground-truth translation of \textit{``sleeping''} and \textit{``slept''} rather than a targeted translation error with respect to $\boldsymbol{r}$. Therefore, both $\boldsymbol{x'}$ are actually invalid, i.e., they do not cause targeted translation errors on the targeted words $w$ since they modify the source sides of the targeted words and break the above assumption. 
%have modified the source sides of the targeted words. 

%Therefore, both $\boldsymbol{x'}$ are invalid adversarial examples, i.e., they actually do not cause targeted translation errors on the targeted words $w$ since they have modified the source sides of the targeted words. 
%have modified the source sides of the targeted words. 

% $\mathcal{M}$ reacts correctly to the modifications in $\boldsymbol{x'}$ and thus we could not claim that $\boldsymbol{x'}$ is a valid adversarial example that leads to translation errors on the corresponding target words. 

% Therefore, many adversarial examples actually do not cause targeted translation errors on target words since they have modified the source sides of these target words.

%In the rest of this section, we empirically quantify the percentage of invalid adversarial examples under these two settings on a standard NMT evaluation dataset.

\vspace{-0.2cm}
\subsection{Empirical Quantification}
\vspace{-0.2cm}

\label{empirical}

%\subsubsection{Setup}
\textbf{Setup.} \quad We experiment on the test set of the WMT20 En-Zh news translation task~\cite{barrault-etal-2020-findings}, which provides 1418 English-Chinese translation pairs. The NMT system we use is a Transformer-based model~\cite{vaswani2017attention} (\textbf{TF}) %with six 512-dimensional encoder and decoder layers, respectively. 
and please refer to \sref{setup} for details %about the training process 
of TF. For both settings, the number of targeted words in each example is set to one.

\textbf{Attack Methods.} \quad
Since attack methods under the same setting share several high-level ideas, we implement the targeted attack \textbf{Seq2sick} designed by~\cite{cheng2020seq2sick} for setting 1, %that searches for token-level perturbations in the continuous embedding space. %to make a specific target word appear in the translation output. 
%As for setting 2, %since the attack methods proposed by~\cite{ebrahimi2018adversarial} and~\cite{wallace2020imitation} share several high-level ideas, 
%we pick 
and the \textbf{Targeted Flips} attack in~\cite{wallace2020imitation} for setting 2 as representatives for further experiments. %Given an input $\boldsymbol{x}$, Targeted Flips iteratively replaces its tokens based on the gradient of the cross entropy between the target word $w$ and other words in the target dictionary to remove $w$ from the translation output. 
All the configuration settings of these two attacks are the same as their original implementations. 
In addition, we implement two more attack methods %as baselines 
under both settings: %for comparison:  
\begin{itemize}[wide=0\parindent,noitemsep,topsep=0em]
\item \textbf{R}andom \textbf{R}eplace (RR): It randomly replaces 30\% of tokens in the input, and its results are averaged over 3 different runs.
\item \textbf{W}hite-box \textbf{TextFooler} (WTextFooler): %A synonym substitution-based adversarial attack 
A word substitution-based attack proposed by~\cite{jin2020bert}, while we use the gradients of the cross entropy between $w$ and other words in the original translation output (Setting 1) / $w$ and words in the target dictionary (Setting 2) to rank the attack positions instead.
\end{itemize}
%Please refer to appendix xxx for more details about the configuration settings of these two attacks methods.

\textbf{Quantification.} \quad
We apply the %NMT targeted adversarial attacks mentioned above 
above attacks to perturb the evaluation set. % and collect all the crafted adversarial examples. 
The effectiveness of different attacks are then measured by the attack success rate (\textbf{Succ}), i.e., the percentage of adversarial examples in all the crafted $\boldsymbol{x'}$, where a higher value refers to better performance. Next, we calculate the percentage of invalid adversarial examples for each attack with the help of an English-Chinese bilingual dictionary\footnote{\url{https://catalog.ldc.upenn.edu/LDC2002L27}}. Given an adversarial example $\boldsymbol{x'}$, \textbf{we consider it as invalid under setting 1 if one of the source-side translations of $w$ in the dictionary appears in $\boldsymbol{x'}$ while $\boldsymbol{x}$ contains none of them, and vice versa under setting 2.} %For instance, the two $\boldsymbol{x'}$ under setting 1 and setting 2 in \fref{fig:compare settings} are invalid since they violate the above rules. 
\tref{tab:empirical quantification}\footnote{We do not attach the results of RR under setting 1 since it fails to generate any adversarial examples.} lists the quantification results. %To our surprise, 
We observe that although most of the attack methods under both settings achieve high attack success rates, the percentages of invalid adversarial examples are nontrivial and make the actual Succ scores of these attacks much lower than the numbers shown in \tref{tab:empirical quantification}. In conclusion, the quantification results convincingly support our argument that the effectiveness of existing NMT targeted adversarial attacks are largely overestimated. 

%Overall, we obtain two findings from the above experiments: %(1): Current settings for NMT targeted adversarial attack will largely overestimate the performances of different attack methods. (2) Attack methods do not work well under existing settings if we filter out invalid adversarial examples. 

%with the help of a English-Chinese bilingual dictionary\footnote{\url{https://catalog.ldc.upenn.edu/LDC2002L27}}. %and a word alignment tool pre-trained on the training corpus, 
%Given an adversarial example $\boldsymbol{x'}$, we consider it as invalid under setting 1 if one of the English translations of $w$ in the dictionary appears in $\boldsymbol{x'}$ while $\boldsymbol{x}$ contains none of them, and vice versa under Setting 2. As an example, the ``Adv In'' for setting 1 and setting 2 in \fref{fig:compare settings} are marked with invalid under ``Human Judgement'' since they violate the above rules.

\begin{table}[tb]
\renewcommand\arraystretch{0.7}
  \centering
   \setlength{\tabcolsep}{3mm}{
    %\resizebox{0.8\textwidth}{!!}
    \small
    \begin{tabular}{cccc}
    \toprule[1pt]
       & \textbf{Method} & \textbf{Succ}%$\uparrow$ 
       & %\textbf{Succ*}$\uparrow$ 
       \textbf{Invalid}%$\uparrow$ 
       \\
       %& \textbf{Edit}$\downarrow$ 
       %& \textbf{PPL}$\downarrow$
       %& \textbf{Query}$\downarrow$
       %& \textbf{\#Invalid}
     
      %\\
      \midrule[0.5pt]
      \multirowcell{2}{\textbf{Setting 1}} 
      %& RR & 0.00 & 0.00 \\
      & WTextFooler & 0.85 %&0.64 
      &25.00\\
      %& & & \\
      ~ & Seq2Sick & 59.17 %&33.00 
      &44.22\\
      %&79.78 &234.57 &44.22 \\
      \midrule[0.5pt]
      \multirowcell{3}{\textbf{Setting 2}} & RR & 53.89 
      %& 20.20 
      & 64.05\\
      %&266.89 &1.00 &64.05 \\
      ~ & WTextFooler &84.64  %&33.14 
      &60.84\\
      %&35.72 &14.01 &60.84 \\
      ~ & Targeted Flips & 45.59 %&15.40 
      &67.71\\
      %& & & \\
     \bottomrule[1pt]
    \end{tabular}
  }  
  %}  
    \caption{Empirical quantification results.%Succ (\%) and Invalid (\%) denote the attack success rate and the percentage of invalid adversarial examples, respectively.}
    }
  \label{tab:empirical quantification}
  \vspace{-0.5cm}
\end{table}

%then build an effective attack method under this new setup.

\section{The Proposed Attack Setting}
\label{framework}

\vspace{-0.2cm}
\subsection{Problem Setup}
\vspace{-0.2cm}

\label{problem setup}
To address the critical issue mentioned before, %in this work, 
we aim to design an appropriate setup for NMT targeted adversarial attacks that could lead to reliable attacking results. %The key idea is to constrain the perturbations on $\boldsymbol{x}$ such that the label-preserving assumption can be held.
%Different from previous works, target words in our proposed evaluation setting are selected from the source input sentences and each evaluation sample is paired with one target word. 
%Following the notations defined in \sref{revisiting},
Given a source sentence $\boldsymbol{x}$ that includes one source side targeted word $z$, %(e.g., \textit{``working''} under our setting in \fref{fig:compare settings})
the goal of an attack under our setting is to craft an adversarial example $\boldsymbol{x'}$ for $\boldsymbol{x}$ that makes the translation of $z$ not appear in the corresponding translation output $\boldsymbol{y'}$, while making sure that $\boldsymbol{x'}$ is still a fluent and meaningful sentence. %which is similar to setting 2 except that our target word is picked from $\boldsymbol{x}$. 
As pointed out by \sref{issue}, the core problem of previous settings is that they fail to hold the label-preserving assumption. %  during the attack process 
%and thus leads to overestimated attacking results. 
To solve this issue, we explicitly constrain that only the non-targeted word tokens in $\boldsymbol{x}$ can be perturbed during the attack process. This step ensures $z$ appears in both $\boldsymbol{x}$ and $\boldsymbol{x'}$, which enables us to find translation errors on $z$ by examining whether the reference translation of $z$ appears in $\boldsymbol{y'}$.
%by comparing $\boldsymbol{y'}$ with the reference of $\boldsymbol{x}$. 

\begin{figure}[tb]
    \centering
    \includegraphics[width=0.44\textwidth]{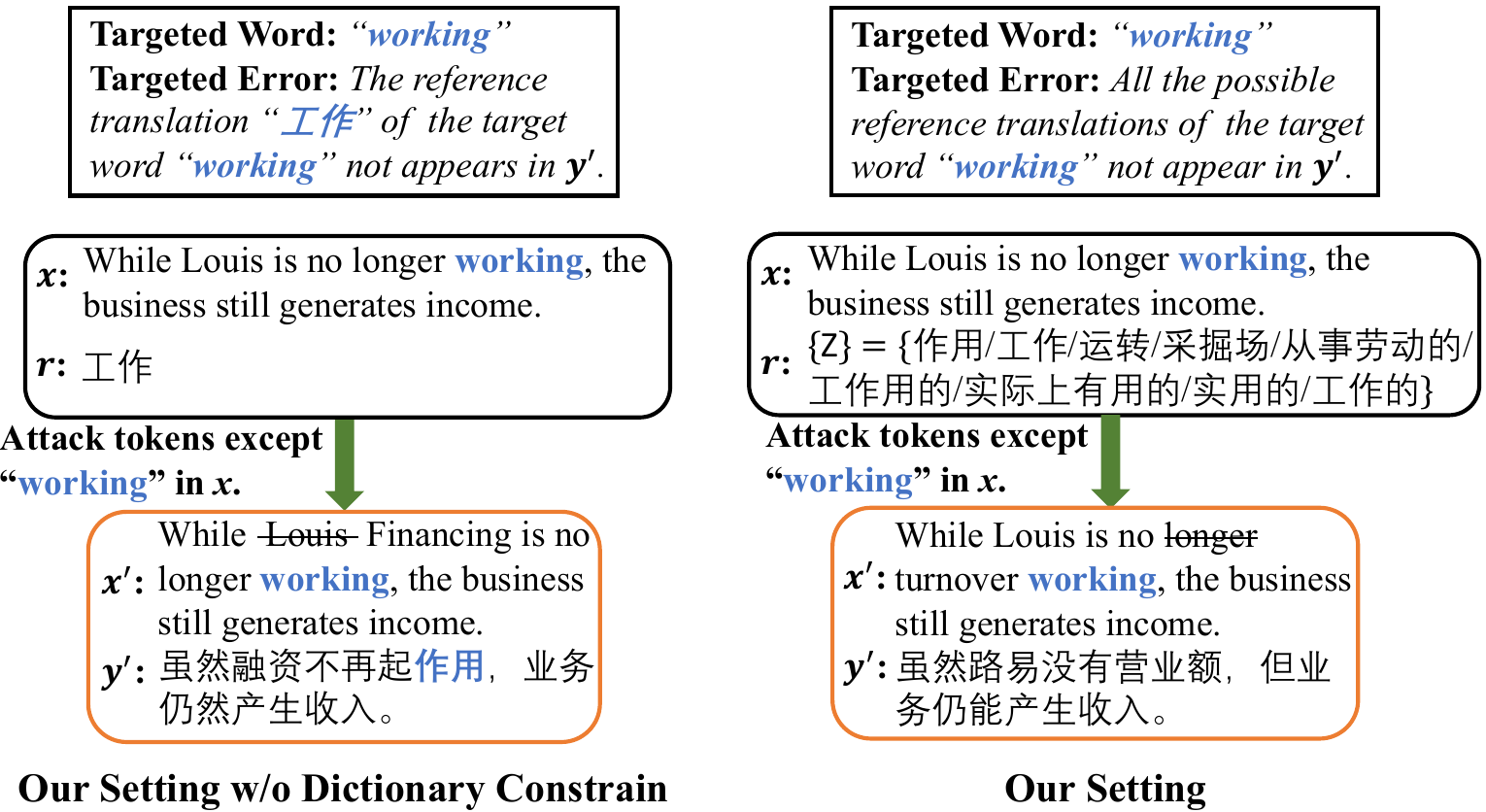}
    \caption{Illustration of our attack setting and its variant.
    %$\boldsymbol{x}$, $\boldsymbol{y}$, $\boldsymbol{x'}$, and $\boldsymbol{y'}$ are the same as \fref{fig:compare settings}. 
    %where $x$ is the original input and $y$ is its reference, $x'$ is the attacked source and $y'$ is its output from the NMT system. %We use the attack proposed by~\cite{cheng2020seq2sick} and~\cite{wallace2020imitation} for Setting 1 and Setting 2, respectively. %Target words and their translations are colored with blue. Modifications are colored with red.
    %Target words and their target side translations are bolded with blue, which indicates the targeted translation errors with respect to the reference translations $\boldsymbol{r}$ of the targeted word.}%the original $\boldsymbol{x}$ and its word-level reference $\boldsymbol{r}$. } 
    }
    \label{fig:our settings}
    \vspace{-1.5em}
\end{figure}

Moreover, since a targeted word $z$ might have various meanings under different contexts, we further apply the bilingual dictionary in \sref{empirical} to obtain a set $\{\mathbf{Z}\}$ that contains multiple reference translations of $z$, and requires a successful attack to make all the translations in $\{\mathbf{Z}\}$ not appear in $\boldsymbol{y'}$.
%(e.g., all the possible Chinese translations for \textit{``working''}
%in \fref{fig:our settings})
%as a word-level reference of $\boldsymbol{x}$
%, which can be used 
%to verify targeted translation errors in $\boldsymbol{y'}$. 
%We regard a targeted attack under our setup as successful only if all the translations in $\{\mathbf{Z}\}$ do not appear in $\boldsymbol{y'}$. 
This step is important since it is possible to craft an invalid adversarial example $\boldsymbol{x'}$ that only makes one translation of $z$ disappear in $\boldsymbol{y'}$. % while letting another appear. 
Consider the left case in \fref{fig:our settings}, the attack is regarded as successful if we only provide one reference translation ``工作'' for the targeted word $z=$ \textit{``working''}.
%since $\boldsymbol{y'}$ makes a targeted translation error on the targeted word \textit{``working''} with respect to the provided reference ``工作''. 
However, since another translation \textit{``作用''} of the targeted word $z=$\textit{``working''} appears in $\boldsymbol{y'}$ and becomes the new ground-truth translation of \textit{``working''},
$\boldsymbol{x'}$ here is actually an invalid adversarial example. 

Finally, we require that the adversarial example $\boldsymbol{x'}$ under our setting should still be a fluent and meaningful sentence, which is important since the semantic meaning of the targeted word $z$ could be destroyed by messy contents in a meaningless sentence. Hence, it could be correct for the reference translation of $z$ not appear in the translation of a meaningless adversarial example $\boldsymbol{x'}$, thus biases the final evaluation results. 

Overall, \textbf{$\boldsymbol{x'}$ is an adversarial example of $\boldsymbol{x}$ in our setting if: 1): it leads to a targeted translation error on the targeted word $z$, i.e., all the reference translations of $z$ in $\{\mathbf{Z}\}$ do not appear in $\boldsymbol{y'}$; 2): it only modifies non-targeted word tokens in $\boldsymbol{x}$. 3): it is a fluent and meaningful sentence.} %Each evaluation sample under our setting can thus be denoted as $\langle \boldsymbol{x}, z,\{\mathbf{Z}\}\rangle$. 
A valid adversarial example under our setting is $\boldsymbol{x'}$ in the right case of \fref{fig:our settings}, where $\boldsymbol{y'}$ makes a targeted translation error on $z=$\textit{``working''}. %with respect to its word-level reference. 
Noting that sentence-level references are not needed in our proposed attack setting, making it easily to be scaled up to provide unbiased attacking results.

\vspace{-0.3cm}
\subsection{Evaluation Metrics}
\vspace{-0.2cm}

%Moreover, we introduce two metrics that are suitable for evaluating an adversarial attack method.
%Given an evaluation set, we apply an NMT targeted adversarial attack to craft adversarial examples for all the evaluation samples and introduce three metrics that are suitable for evaluating this attack method.
% for $\mathcal{M}$,
We introduce three metrics that are suitable for evaluating an NMT targeted adversarial attack method.
First, %given an evaluation set, we apply an NMT targeted adversarial attack to craft adversarial examples for all the evaluation examples and 
we compute the attack success rate (\textbf{Succ}) defined in \sref{revisiting} 
%i.e., the percentage of evaluation examples that their adversarial examples can be constructed according to a certain attack method, 
to measure the effectiveness of the given attack method, where a higher Succ score means it is more powerful. %In addition, we measure the named entity attack success rate (\textbf{NSucc}), i.e., the proportion of incorrectly translating a given named entity $n$ in all examples crafted by an attack. 
%The above two metrics are used to evaluate a model's adversarial robustness performance in the entity-targeted translation setting.
Second, we assume that a good attack method should generate adversarial examples with few token modifications, which avoids it from being a messy sentence.
Therefore, we compute %the average token modification percentage on the generated adversarial examples,which is 
the edit distance (\textbf{Edit}) between $\boldsymbol{x'}$ and $\boldsymbol{x}$, normalized by the sequence length of $\boldsymbol{x}$. 
%Second, a good attack method should generate its adversarial examples as human-like sentences. Hence, we calculate the perplexity (\textbf{PPL}) using a large-sized GPT-2 language model \cite{radford2019language}. 
Finally, we suppose a good attack method should craft adversarial examples efficiently and thus record the average number of times an attack queries the victim NMT system for generating an adversarial example (\textbf{Query)}. For the latter two metrics, a lower value refers to better performance.
\vspace{-1.0em}

\section{TWGA Method}
\label{generate adv}
Next, we propose a novel white-box \textbf{T}argeted \textbf{W}ord \textbf{G}radient adversarial \textbf{A}ttack (\textbf{TWGA}) method to craft effective adversarial examples for NMT systems under our attack setting. We get inspired by \cite{guo2021gradient}, in which they propose to generate token modification-based adversarial examples with a parameterized distribution. 
%Specifically, TWGA aims to make the correct translations of $\boldsymbol{x}$'s targeted word $z$ disappear from the corresponding translation output by subtly modifying non-targeted word tokens in $\boldsymbol{x}$, as illustrated in our setting in \fref{fig:our settings}.
However, the method in \cite{guo2021gradient} is designed for text classification, where each input only has one label. In our work, we try to search for an ideal distribution of adversarial examples in a sequence prediction task with an ultra high-dimensional output space. To this end, we propose to record the output logits of $\mathcal{M}$ at each decoding step, 
which helps to marginalize the probability vectors of $z$ for crafting better adversarial examples targeting it. %The algorithm of TWGA is shown in Algorithm~\ref{alg:adversarial example generation} in Appendix, 
We will explain its steps in detail in the following.

\textbf{Initializing the Probability Matrix.} \quad
The source input $\boldsymbol{x}$ is first tokenized into sub-word tokens $(x_{1},\cdots, x_{i},\cdots, x_{n})$. %by the Byte-Pair Encoding (BPE) scheme~\cite{sennrich2015neural}. 
Suppose $x_i$ 
has an index
$j$ in the source vocabulary $\mathbf{V}^{src}$. 
We then initialize $\mathbf{P} \in \mathbb{R}^{n \times |\mathbf{V}^{src}|} $ from $\boldsymbol{x}$ by 
setting all values to zero except $\mathbf{P}_{i,j}=\epsilon$, where $\epsilon$ is a constant.
Next, we apply the Gumbel-softmax strategy~\cite{jang2016categorical}
to approximate a differentiable distribution $\Gamma$ 
upon $\mathbf{P}$:

\vspace{-0.2cm}
\begin{equation}
\label{gumbel}
    \Gamma_{i,j} = \frac{\exp((\mathbf{P}_{i,j} + g_{i,j})/\tau)}{\sum_{v=1}^{|\mathbf{V}^{src}|}\exp((\mathbf{P}_{i,v} + g_{i,v})/\tau)}
\end{equation}
\vspace{-0.2cm}

where $g_{i,j}$ is randomly drawn from $\text{Gumbel}(0,1)$, which enables us to obtain different $\Gamma$ with a fixed $\mathbf{P}$. $\tau$ is a temperature parameter. This approximation 
step enables valuable gradient information provided by our objective function to assist the optimization of $\mathbf{P}$.

\textbf{Objective Function.} \quad
\label{para:objective}
We also tokenize the reference translations of the target word $\{\mathbf{Z}\}$ %using the BPE scheme 
and put the corresponding sub-word tokens into $\{h_p\}$.
%to form $\{h_p\}$.
\iffalse
\footnote{
 Two types of tokens % in the target vocabulary  
 will also be added to $\{h_p\}$: 
%\begin{enumerate}[wide=0\parindent,noitemsep,topsep=0em]
%\setlength{\itemsep}{0.5pt}
%\setlength{\parsep}{0.5pt}
%\setlength{\parskip}{0.5pt}
    (i) $h_{p}$ 's counterpart ``$h_p$@@'' created by BPE;
    (ii) Tokens that incorporate $h_{p}$ (e.g., \textit{``工作的"} incorporates \textit{``工作''}).}
\fi
Suppose we apply beam search as the decoding method (yet other decoding strategies that rely on $\mathcal{M}$'s output logit vector to pick tokens
can also be used). 
At each decoding step, we record $\mathcal{M}$'s output logit vector $\Theta$ of the specific beam
and use a margin loss $\mathcal{L}_{\text{mar}}$ that encourages tokens in $\{h_p\}$ not to appear in the translation output 
to optimize $\mathbf{P}$:

\vspace{-0.4cm}
\begin{equation}
\label{margin used}
    \mathcal{L}_{\text{mar}} = 
    %\mathcal{L} = 
    \sum_{h \in \{h_p\}}
    %\!\sum_{q=1}^{len(\boldsymbol{y'})}\!
    \max \! \left(\!\Theta_{{h}}- \max\limits_{t \in \textbf{V}^{tgt} /\{h_p\}}\!\Theta_{t} + \mu, 0 \! \right) \nonumber
    \end{equation}
\vspace{-0.4cm}

where $t$ is a token in the target vocabulary $\textbf{V}^{tgt}$ except the tokens in $\{h_p\}$.
The value of $\mathcal{L}_{\text{mar}}$ is bounded by the decision margin $\mu$ when $h$ is not the Top-1 prediction at this step. 
$\Theta_{{h}}$ and $\Theta_{t}$ are the logit values for $h$ and $t$ in $\Theta$, respectively.
We then sum $\mathcal{L}_{\text{mar}}$ for all decoding steps, denoted as $\mathcal{L}_{\text{adv}}$. %This optimization process aims to make the approximated distribution $\Gamma$ become an adversarial distribution.

Besides, we add two loss functions to regularize that the edited input sequences still approximate fluent and meaningful sentences. %to fulfill the requirement of our attack setting. 
We first sample the word index from each row of $\Gamma$ to obtain the edited sequence, then minimize the negative log likelihood of this sequence using two pre-trained language models with the %small-sized 
 GPT-2's architecture~\cite{radford2019language}. One is pre-trained to generate left-to-right language modeling probabilities, and the other one provides right-to-left probabilities. %\footnote{See Appendix~\ref{appendix: language models} for training details.}. 
 Denote these two losses as $\mathcal{L}_{\text{nll}}$ and $\mathcal{L}_{\text{nll}'}$, 
%The two language models are with the small-sized GPT-2's architecture~\cite{radford2019language}\footnote{See Appendix~\ref{appendix: language models} for training details.}. 
our objective function can be written as:

\vspace{-0.2cm}
\begin{equation}
\label{new adv objective}
    \min\limits_{\mathbf{P}}\ \mathbb{E}_{\Gamma \sim \mathbf{G}}[\frac{1}{k}\mathcal{L}_{\text{adv}} +\lambda_{1} \mathcal{L}_{\text{nll}} + \lambda_{2} \mathcal{L}_{\text{nll}'}]
    %\end{aligned} 
    %\nonumber
\end{equation}
\vspace{-0.2cm}

%\bi{what is G}
where $\lambda_{1}$ and $\lambda_{2}$ are two hyper-parameters and $k$ is the number of translations in $\{\mathbf{Z}\}$. $\mathbf{G}$ refers to the Gumbel-softmax distribution in Eq.\ref{gumbel}.  
%We then use this loss %~(\ref{new adv objective}) 
%in Eq.\ref{new adv objective} to optimize %the distribution 
%$\mathbf{P}$.
$\mathbf{P}$ is then optimized by minimizing the above objective via gradient descent.
Note that if the sub-word tokenized sequence of the targeted word $z$ includes $x_i$, $\mathbf{P}_{i}$ 
will be fixed during the optimization since TWGA only modifies non-targeted word tokens.

\textbf{Crafting Adversarial Examples.} \quad
%With the optimized $\mathbf{P}$, we show how to craft an adversarial example for $\boldsymbol{x}$.
When crafting adversarial examples, we first obtain a distribution $\Gamma$ on top of the optimized $\mathbf{P}$ via Eq.\ref{gumbel}. Then, 
for each non-targeted word token $x_i \in \boldsymbol{x}$, we replace it with another token
that has the highest probability in $\Gamma_i$ to create $\boldsymbol{x'}$. Note that to further constrain the search space between $\boldsymbol{x}$ and $\boldsymbol{x'}$, the new token 
%while fulfilling the following requirement: this token 
and $x_i$ should be both with/without the BPE signal ``@@'' and cased/uncased.
If such token does not exist, $x_i$ will remain static. %Afterward, we get a modification $\boldsymbol{x'}$ of $\boldsymbol{x}$. 
If $\boldsymbol{x'}$ obtained from the above process 
%meets Definition \ref{def2}, 
makes a targeted translation error on the targeted word $z$ with respect to $\{\mathbf{Z}\}$,
it becomes our final adversarial example for $\boldsymbol{x}$.
%Recall that we can get various distribution $\Gamma$ from the optimized $\mathbf{P}$ by sampling different $g_{i,j}$ to calculate Eq.\ref{gumbel}. % for multiple times. Hence, 
We repeat the above process for multiple times via sampling different $g_{i,j}$ to calculate Eq.1 until an adversarial example $\boldsymbol{x'}$ appears. We try 100 times at most for each input $\boldsymbol{x}$. Otherwise, we consider that TWGA fails to construct an adversarial example for $\boldsymbol{x}$. 
\begin{table}[tb]
\renewcommand\arraystretch{0.7}
  \centering
   \setlength{\tabcolsep}{2mm}{
    %\resizebox{0.8\textwidth}{!!}
    \small
    \begin{tabular}{cccc}
    \toprule[1pt]
    &
      \textbf{Evaluation Set} & \textbf{\#Example} 
      & \textbf{Avg.Length}
      \\
     \midrule[0.5pt]
     LSTM &WMT & 2311&36.67   
     \\
     
     \midrule[0.5pt]
     \multirowcell{2}{$\text{TF}$} &
      WMT & 2263&36.16  \\
      ~&Para & 53834 & 22.94   \\
     \bottomrule[1pt]
    \end{tabular}
  }  
  %}  
    \caption{Statistics of the constructed evaluation sets.}% Avg.Len denotes the averaged length of source sequences in an evaluation set.}
  \label{tab:statistics}
\vspace{-0.5cm}
\end{table}

\section{Experiments}
\label{exp}
\iffalse
%In this paper, 
%First, we demonstrate the importance of our proposed setting on NMT targeted adversarial robustness evaluation.
First, we demonstrate the effectiveness of our proposed attack method TWGA on two NMT systems by comparing it with several baseline attacks under our attack setup, and conduct a human evaluation to investigate the meaningfulness of adversarial examples crafted by TWGA.
%Next, we use TWGA to comprehensively evaluate the adversarial robustness of two prevailing NMT systems. 
Next, we analyze the possible bias in the attacking results caused by limited evaluation data using a large-scale dataset. Finally, we illustrate several insightful findings through further investigation on the large-scale dataset. %Our code and data will be released on http://github.xxx.
We will release our code and data after the anonymized review period.
\fi

\vspace{-0.2cm}
\subsection{Settings}
\vspace{-0.2cm}

\label{exp setup}
\textbf{Evaluation Set.} \quad
We conduct all experiments on the English-Chinese translation task and construct a raw evaluation set \textbf{WMT} using the English data in the test set of WMT20's En-Zh news translation task. %to %measure the adversarial robustness of NMT systems evaluate different NMT targeted adversarial attacksunder our proposed setting. 
Given a source sentence $\boldsymbol{x}$, we first use NLTK %~\footnote{\url{https://www.nltk.org/}} 
to find all its nouns, verbs, adjectives, and adverbs, then filter out stopwords and words that do not appear in the bilingual dictionary mentioned in \sref{empirical} to form a set of targeted words candidates. If the candidate set has more than three words, we randomly sample three as targeted words to create three separate evaluation samples $\langle \boldsymbol{x}, z,\{\mathbf{Z}\}\rangle$. Otherwise, all the words in the candidate set will be used as targeted words to create individual evaluation samples. For each victim NMT system, we build its evaluation set on top of the raw evaluation set by picking evaluation samples whose targeted word $z$ is translated correctly with respect to $\mathbf{Z}$.
%its correctly translated sentences according to Definition \ref{def1}. 
The evaluation set statistics are shown in \tref{tab:statistics}.

\textbf{NMT Systems.} \quad
\label{setup}
 Two prevailing NMT systems are used in our experiments. The first is an LSTM-based model \cite{luong2015effective} (\textbf{LSTM}) %that contains an encoder layer and a decoder layer with 512 hidden units. 
and the other one is the TF translation model introduced in \sref{empirical}.
For both systems, the word-embedding size is set to 512. We train the NMT systems using the WMT20 En-Zh news translation corpus that includes 31M sentence pairs~\cite{barrault-etal-2020-findings} for 300K steps. During training, we use the Adam optimizer ($\beta_1=0.9, \beta_2=0.98)$ with a learning rate of 1e-4. Dropout rate is set to 0.3 and label smoothing rate is set to 0.2. Two separate BPE models are applied to generate vocabularies for English ($\sim$48K tokens) and Chinese ($\sim$59K tokens)%, respectively
These two NMT models achieve an average BLEU score of 27.49 and 38.66 on the WMT20 En-Zh test set %\footnote{For reference, the rank 1st system~\cite{shi2020oppo} achieves an BLEU score of 43.20.}, 
respectively. 

\textbf{Baseline Targeted Adversarial Attack Methods.} \quad
To demonstrate the effectiveness of TWGA, we implement %the four targeted attacks 
RR, WTextFooler(Setting 2), Targeted Flips, and Seq2Sick which are described in \sref{empirical} as baselines under our proposed attack setting. 
We do not include the attack method proposed by~\cite{sadrizadeh2023targeted} since it applies the cross-entropy loss as the loss function, rather than the hinge-like loss used by Seq2Sick that is more suitable for our proposed setting.
For WTextFooler and Targeted Flips, we %attack a sample $\langle \boldsymbol{x}, z,\ {\mathbf{Z}\}\rangle$ by making 
regard $z$'s translation that appears in $\boldsymbol{y}$ as their required targeted word $w$. 
As for Seq2Sick, we negate its objective function since we want the correct translation of $z$ to be removed from the original translation output. Noting that we only allow these attacks to modify non-targeted word tokens in the source sentence $\boldsymbol{x}$ as required in our proposed attack setup.

%We implement RR, WTextFooler (Setting 2), and Targeted Flips as described in \sref{empirical}, while negating the objective function of Seq2Sick since we want the correct translation of the target word to disappear from the translation output. Noting that we only allow these attack methods to modify non target word-tokens in the source sentence as required in our proposed setup.

%\begin{itemize}[wide=0\parindent,noitemsep,topsep=0em]
%\item \textbf{R}andom \textbf{R}eplace (RR): It randomly replaces 30\% tokens except the target word in the input sentence and its results are averaged over three different runs.
%\item \textbf{Targeted Flips} (Flip): A white-box targeted attack proposed by~\cite{wallace2020imitation}. 
%\item \textbf{Seq2sick}: The targeted attack used in~\cite{cheng2020seq2sick}, while we negate its objective function since we want the correct translation of the target word to disappear. 
%For the core ideas and more details about the above two attack baselines, please refer to \sref{empirical} and appendix XXX.
%\end{itemize}

\begin{table}[tb]
\renewcommand\arraystretch{0.7}
  \centering
   \setlength{\tabcolsep}{2.2mm}{
    %\resizebox{0.8\textwidth}{!!}
    \small
    \begin{tabular}{llccc}
    \toprule[1pt]
    &
      \textbf{Method} & \textbf{Succ}$\uparrow$ 
      & \textbf{Edit} $\downarrow$ 
      %&  \textbf{PPL} $\downarrow$ 
      & \textbf{Query} $\downarrow$ 
      \\
     \midrule[0.5pt]
     \multirowcell{5}{$\text{LSTM}$} &
     RR & 14.85
     & 32.00 %& 310.29 
     & -
     \\
     ~ & WTextFooler & 45.05
      & 7.09 %& 38.51
      & 42.68
      \\
       ~ & Targeted Flips & 33.85
      & \textbf{5.31} %& 70.21 
      & 719.27
      \\
       ~ & Seq2sick & 57.68
      & 14.21 %& 78.14 
      & 226.71
     \\
     ~& TWGA &  \textbf{82.82}
    & 12.15 %& 95.57 
    & \textbf{36.97} \\
     
     \midrule[0.5pt]
     \multirowcell{5}{$\text{TF}$} &
      RR & 10.60
     & 32.05 %& 285.11 
     & -
     \\
     ~ & WTextFooler & 32.79
      & 7.82 %& 39.49 
      & 49.79
      \\
       ~& Targeted Flips & 34.47
      & \textbf{6.85} %& 69.21 
      & 1466.39
      \\
       ~& Seq2sick &47.19
      & 14.62 %& 107.82 
      & 263.86
     \\
     ~& TWGA &  \textbf{72.91}
    & 12.15  %& 83.89 
    & \textbf{29.63}\\
     \bottomrule[1pt]
    \end{tabular}
  }  
  %}  
    \caption{Results of different attacks on WMT. Succ/Edit are reported by percentage (\%).
     The best performances on each set are boldfaced.}
  \label{tab:attack comparison}
  \vspace{-0.5cm}
\end{table}

\vspace{-0.3cm}
\subsection{Results of Different Attack Methods}
\vspace{-0.2cm}

\label{attack effectiveness evaluation}
%\paragraph{Automatic Evaluation Results.} 
As shown in \tref{tab:attack comparison}\footnote{The Query scores of RR are not shown since they are always 1.00.}, 
%all the other four attack methods achieve better performances than RR, showing that simply perturbing tokens without a proper strategy will not seriously affect targeted word translation. 
among these four attack methods, TWGA has significantly higher Succ scores on both LSTM and TF, demonstrating that the optimized adversarial distribution used in TWGA is more suitable for searching optimal adversarial examples. 

%As for the semantic measurements, %TWGA still outperform RR and Seq2sick. 
Although WTextFooler and Targeted Flips have lower Edit scores than TWGA since they apply several heuristic rules to constrain the token substitution processes, their perturbation spaces are also limited by these rules and results in poor performances on Succ and Query. 
In contrast, TWGA achieves better Succ and Query results while having comparable Edit scores with the above two attacks, hence attaining a good trade-off on all the evaluation metrics. 
%\vspace{-1.5em}
%demonstrating its better ability to balance the attacking effectiveness, efficiency and quality by simply adjusting the hyperparameters in Eq.\ref{new adv objective}. 
%More importantly, the fact that TWGA needs very few queries to craft adversarial examples enables it to be applied on large-scale evaluation sets, which further extends the application scenarios of TWGA.

\begin{table}[tb]
\renewcommand\arraystretch{0.7}
  \centering
   \setlength{\tabcolsep}{0.7mm}{
    \small
    \begin{tabular}{lcccc}
    \toprule[1pt]
    \textbf{Method}
    & \textbf{1-meaningless} 
      & \textbf{2-mediocre} 
      & \textbf{3-meaningful} 
      \\
     \midrule[0.5pt]

    WTextFooler & 0.13 & 0.57 & 0.30 \\
    Targeted Flips & 0.08 & \textbf{0.66} & 0.26 \\
    Seq2sick & \textbf{0.16} & 0.63 & 0.21 \\
    TWGA & 0.06 & 0.60 & \textbf{0.34} \\
     \bottomrule[1pt]
    \end{tabular}
  }  
    \caption{Score distributions of human evaluation on adversarial examples crafted by different attack methods.}
  \label{tab:human}
  %\vspace{-1.0em}
\end{table}

\begin{table}[tb]
\vspace{-1.0em}
\renewcommand\arraystretch{0.7}
  \centering
   \setlength{\tabcolsep}{3mm}{
    %\resizebox{0.8\textwidth}{!!}
    \small
    \begin{tabular}{lcccc}
    \toprule[1pt]
    \textbf{Method} 
    & \textbf{Succ}$\uparrow$ 
      & \textbf{Edit} $\downarrow$ 
      %&  \textbf{PPL} $\downarrow$ 
      & \textbf{Query} $\downarrow$ 
      \\
     \midrule[0.5pt]
     RR & 8.24 & 34.19 %&  
     & - \\
     Seq2sick & 42.44 & 26.18 %&  
     & 51.62 \\
     TWGA & \textbf{57.87} & \textbf{13.76}  %& 163.92 
     & \textbf{33.92} \\
     \bottomrule[1pt]
    \end{tabular}
  }  
    \caption{Attack results of TF on Para. Succ/Edit are reported by \%.}
  \label{tab:large scale attacking results}
  \vspace{-0.5cm}
\end{table}

\vspace{-0.3cm}
\subsection{Meaningfulness of Adversarial Examples}
\vspace{-0.2cm}

\label{meaningfulness}
Since adversarial examples crafted under our setting should be fluent and meaningful sentences, except measuring the Edit score in~\tref{tab:attack comparison}, we further perform a human evaluation on the meaningfulness of adversarial examples generated by different attack methods. Specifically, we first randomly pick 100 sentences from TF's evaluation set on WMT that are successfully attacked by all the attack methods, then score the meaningfulness of adversarial examples crafted by different attacks following a 3-point scale.%~\footnote{See Appendix~\ref{appendix:human} for instructions of this scoring scale.}.

Results are listed in~\tref{tab:human}, where we have two conclusions. First, most adversarial examples generated by TWGA are not meaningless (94\%), which further ensuring the faithfulness of its evaluation results. Also, TWGA generates more meaningful adversarial examples than other baselines, which again demonstrates its strong ability.

\vspace{-0.3cm}
\subsection{Analysis on a Large-scale Evaluation Set}
\vspace{-0.2cm}

\label{large scale analysis}
As mentioned in \sref{introduction}, one advantage of our proposed attack setting is that it does not require sentence-level references and thus can be easily scaled up to provide unbiased results. In this subsection, we further evaluate NMT targeted adversarial attacks on a large-scale dataset for the first time using TF.

\textbf{Large-scale Evaluation Set.} \quad
We construct a large-scale raw evaluation set \textbf{Para} using monolingual data from the open-sourced Paracrawl corpora~\cite{banon-etal-2020-paracrawl}. To avoid overlapping with the training data of the victim NMT model, we take the English side of Paracrawl's English-French corpus, rather than English-Chinese, as our monolingual data. The following steps for building Para are the same as WMT, except that we stop this process when the number of the source sentences $\boldsymbol{x}$ reaches 30K. Statistics of Para are also shown in~\tref{tab:statistics}. %As can be seen, the evaluation set on Para is much larger than WMT, thus supporting us in comparing the attacking results of an NMT system on different data scales.

\textbf{Attacking Results on Para.} \quad
\label{attack results on para}
Except for TWGA, we also implement RR and Seq2Sick for comparison. %, while we reduce the Query value of Seq2Sick to enable it to be applied on the large set\footnote{The implementation details are provided in Appendix~\ref{appendix:seq2sick}.}. %, where we show that this Query-reduced version of Seq2Sick serves as a good approximation of the original implementation.}. 
\tref{tab:large scale attacking results} shows the attacking results of TF on its evaluation set on Para. We find that similar to the results obtained on WMT, TWGA outperforms the other two baselines across the board, which again illustrates its superiority in crafting adversarial examples. The large scale of Para further makes this comclusion more unbiased.

\section{Conclusion}
In this work, we indicate a critical limitation in the existing settings of NMT targeted adversarial attacks that could lead to unreliable attacking results. To this end, we propose a new setting for NMT targeted adversarial attack  %named entities for reliable evaluation results. 
%Under the new setting,
to obtain faithful attacking results. Under the new setting, we design an effective attack TWGA to craft adversarial examples, whose strength is demonstrated by extensive experiments. 
Experiments on a large dataset further lead to some insightful findings about NMT targeted adversarial attacks. 
\bibliographystyle{IEEEbib}
\bibliography{strings,refs}

\end{CJK*}
\end{document}